\documentclass[letterpaper, 10 pt, conference]{ieeeconf}  

\IEEEoverridecommandlockouts                              

\usepackage{algorithmic}
\usepackage{subcaption}
\usepackage{graphicx} 
\usepackage{cite}
\usepackage{caption}
\usepackage{amsmath,amssymb,amsfonts}
\usepackage{makecell}                 
\usepackage{booktabs}                 

\title{\LARGE \bf
YO-CSA-T: A Real-time Badminton Tracking System Utilizing YOLO Based on Contextual and Spatial Attention
}

\author{Yuan Lai$^{1}$, Zhiwei Shi$^{1}$ and Chengxi Zhu$^{2}$
\thanks{$^{1}$ Authors are with School of Computer Science and Technology, Shandong University, Qingdao, Shandong 266237, China
    {\tt\small
    wtttcl@mail.sdu.edu.cn
    }}%
}

\begin{document}

\maketitle
\thispagestyle{empty}
\pagestyle{empty}

\begin{abstract}

The 3D trajectory of a shuttlecock required for a badminton rally robot for human-robot competition demands real-time performance with high accuracy. However, the fast flight speed of the shuttlecock, along with various visual effects, and its tendency to blend with environmental elements, such as court lines and lighting, present challenges for rapid and accurate 2D detection. In this paper, we first propose the YO-CSA detection network, which optimizes and reconfigures the YOLOv8s model's backbone, neck, and head by incorporating contextual and spatial attention mechanisms to enhance model's ability in extracting and integrating both global and local features. Next, we integrate three major sub-tasks—detection, prediction, and compensation—into a real-time 3D shuttlecock trajectory detection system. Specifically, our system maps the 2D coordinate sequence extracted by YO-CSA into 3D space using stereo vision, then predicts the future 3D coordinates based on historical information, and re-projects them onto the left and right views to update the position constraints for 2D detection. Additionally, our system includes a compensation module to fill in missing intermediate frames, ensuring a more complete trajectory. We conduct extensive experiments on our own dataset to evaluate both YO-CSA's performance and system effectiveness. Experimental results show that YO-CSA achieves a high accuracy of 90.43\% mAP@0.75, surpassing both YOLOv8s and YOLO11s. Our system performs excellently, maintaining a speed of over 130 fps across 12 test sequences.

\end{abstract}

\section{INTRODUCTION}

Over the past decade, deep learning has advanced rapidly and has found widespread applications across numerous fields, sparking  multiple high-profile human-machine competitions. 

However, in the field of badminton match, existing research still faces many challenges before achieving realistic human-machine competitions, including real-time extraction of shuttlecock's 3D trajectory, prediction of its future landing point, formulation, and timely optimization of robotic arm movements and striking strategies. Among these, real-time extraction of shuttlecock's 3D trajectory, as the first step in human-machine competitions, directly impacts the effectiveness of subsequent strategies due to its speed and accuracy.

Among various small ball sports, unlike spherical structures such as tennis and table tennis, shuttlecock has a more complex conical structure composed of a cork base and feathers. During high-speed flight, the shuttlecock's shape in images varies significantly depending on the viewing angle. Additionally, during rallies, its high-speed motion can cause the shuttlecock to blend seamlessly with background elements, such as court lines, or lighting. This makes even the 2D detection of the shuttlecock quite challenging.

Accurately and in real time capturing shuttlecock's trajectory in 3D space not only provides valuable data for predicting its landing point and formulating a striking strategy for robots, but also enables match data analysis. This can facilitate the creation of 3D-based automated match analysis systems, helping coaches precisely assess players' capabilities and develop more focused and efficient training plans for athletes.

Against this backdrop, we aim to propose a 3D shuttlecock tracking system that integrates real-time performance with high accuracy. Considering the shuttlecock's small size and susceptibility to false detection, we propose a 3D shuttlecock tracking system based on YO-CSA detection network. We build a simplified stereo vision system to map 2D trajectories to 3D space. Additionally, our tracking algorithm integrates three key modules, shuttlecock detection, prediction, and compensation, to achieve robust and reliable tracking.

To validate the effectiveness of our system, we compare our detection module against mainstream networks. The results show that YO-CSA detection network significantly outperforms these baseline models, particularly in the mAP@0.75 metric. Additionally, we conduct comparative experiments on our tracking strategy which demonstrate the system's effectiveness.

\section{Related Work}

\subsection{Object Detection}

Object detection has extensive applications, including the identification of nearby pedestrians and vehicles in autonomous driving, and intelligent target tracking in surveillance systems. Similarly, object detection technology is crucial for badminton rally robots. By providing the coordinates of the shuttlecock, object detection methods enable the robot to predict the future landing location of the shuttlecock, thereby optimizing its movement and return strategies.

Beginning with R-CNN\cite{girshick2014rich}, which achieving a mean average precision (mAP) of 53.3\% on VOC12\cite{everingham2015pascal}, object detection has rapidly advanced over the past decade. In 2016, the introduction of YOLO\cite{redmon2016you} marks a significant breakthrough, establishing the single-stage model paradigm. This approach detects objects directly on the entire image, and can simultaneously predict object locations and class probabilities in a single inference. More recently, popular backbone networks like CenterNet\cite{duan2019centernet} and EfficientNet\cite{tan2019efficientnet}, along with Transformer-based models such as DETR\cite{zhu2020deformable} and Swin Transformer\cite{liu2021swin}, gain widespread use. The latest YOLO models, including YOLOv8 and YOLO11, continue to push the limits of performance in the field.

To address the challenges posed by the high-speed movement of the shuttlecock, we design the YOLO based on contextual and spatial attention (YO-CSA). This network is built on YOLOv8s, a model known for its high accuracy and fast inference speed, to effectively handle the detection of the shuttlecock under such conditions.

\subsection{Self-Attention Mechanism}

The self-attention mechanism, first introduced in\cite{vaswani2017attention}, differs from traditional convolution operations, which process the entire image uniformly. Self-attention allows the network to dynamically adjust its focus on specific regions, enabling it to prioritize target objects while reducing attention to less important areas, such as the back-ground. It also facilitates the capture of long-range dependencies, overcoming the locality limitations of the convolutional receptive field.

Originally developed for natural language processing tasks, the self-attention mechanism has shown promising results in the field of computer vision as well. ViT\cite{dosovitskiy2020image} is the first to apply Transformer architecture to image recognition, achieving an accuracy of 88.55\% on ImageNet\cite{deng2009imagenet}, even surpassing ResNet\cite{he2016deep}. \cite{dai2023td}\cite{yang2023tse} demonstrate the practical use of Transformers in medical imaging, where they effectively leverage self-attention to capture global features. Additionally,\cite{wang2023swin}, based on ViT and fully connected conditional random fields, optimize the modeling of point cloud data. In object detection, DETR\cite{zhu2020deformable} entirely eliminates the region proposal, region extraction, and non-maximum suppression steps, achieving end-to-end object detection using a pure Transformer architecture. Swin Transformer\cite{liu2021swin} further addresses the computational inefficiencies of ViT by introducing local windows for self-attention.

While self-attention mechanisms can greatly improve network performance, pure Transformer-based models typically require large datasets for training, and their high complexity can limit inference speed—factors that do not align with our objectives. Consequently, we focus on combining the advantages of both convolution and self-attention mechanisms to enable real-time detection of the shuttlecock.

\subsection{Tracking in Small-Size Ball Sports}

Small ball sports, such as badminton and table tennis, are globally popular and have attracted significant research interest. Whether as essential data for human-machine competitions or as part of a match video analysis system, extracting the ball's trajectory is crucial.

CenterNet detects targets by learning a center heatmap. Similarly, a series of works adapt the same fundamental approach by using heatmaps for precise localization. TTNet\cite{voeikov2020ttnet} is a multifunctional neural network designed for high-resolution video, integrating table tennis ball detection, event classification, and semantic segmentation. It takes consecutive video frames as input, using a pure convolution network and heatmap learning, performs coarse-to-fine detection and segmentation of the ball. TrackNetV2\cite{sun2020tracknetv2}, inspired by the UNet\cite{ronneberger2015u}, introduces heatmaps and uses a Gaussian 2D distribution to determine the ball's location. WASB-SBDT\cite{tarashima2023widely} builds a neural network to predict the heatmap of ball coordinates, inspired by HRNet\cite{wang2020deep}. It proposes High-Resolution Modules (HRMs) to address the semantic and spatial resolution loss typically seen in traditional encoder-decoder architectures.

On the other hand, some works focus on optimizing existing detection frameworks. For example, \cite{wu2022deep} enhances model training using gradient estimation theory, improving tennis ball recognition accuracy in videos. \cite{zhang2020efficient} combines YOLOv3\cite{redmon2018yolov3} with Kalman filter\cite{bishop2001introduction} to achieve a 2D golf ball tracking system. \cite{li2023table} utilizes Transformer-based secondary feature processing to build global information, proposing a target detection algorithm for identifying the table tennis ball and determining its position on the table.

Most current research focuses on 2D video processing, with few addressing trajectory extraction in 3D space. This is likely due to the high cost of building devices capable of supporting 3D trajectory extraction. Furthermore, video analysis tasks, unlike robot game-play, do not require 3D data.The field of robotics is still developing, and current research is relatively limited. Therefore, this paper focuses on the development of a real-time, accurate 3D badminton tracking system, aiming to make a meaningful contribution to the field of robotics.

\section{Real-time Detection Module}

In this section, we introduce a novel detection network integrated contextual and spatial attention mechanisms as well as convolution to balance the trade-off between speed and accuracy in high-speed, small-object detection for badminton.

\begin{figure}[h]
    \centering  
    \includegraphics[width=0.48\textwidth]{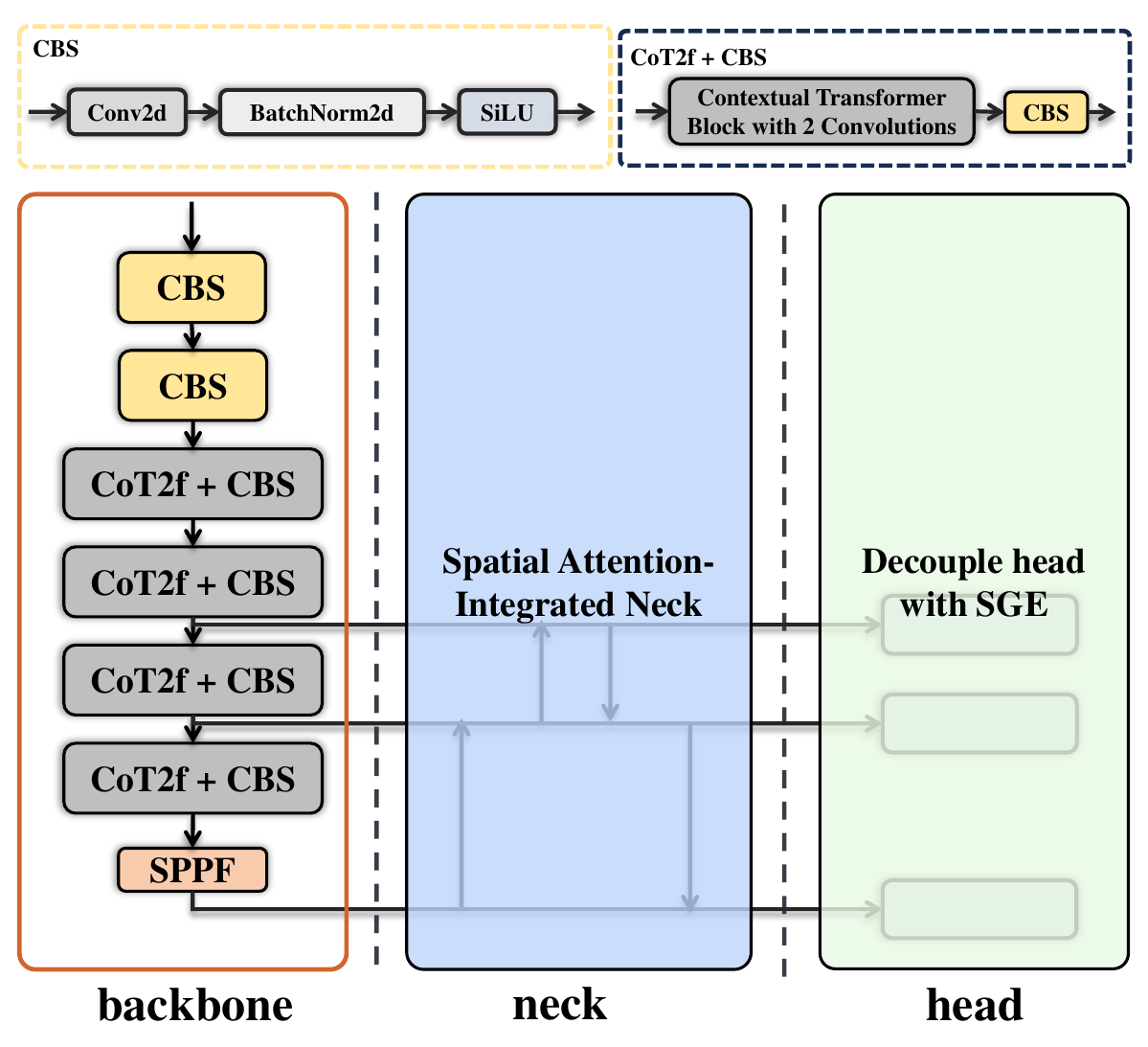}
    \caption{Structure of YO-CSA}
    \label{fig:yocsa}  
\end{figure}

\subsection{Brief Review of YOLO}\label{AA}
Since the introduction of YOLO\cite{redmon2016you} in 2016, the YOLO series has undergone annual iterations, consistently leading the field of object detection with its superior performance. As an end-to-end architecture, YOLO stands in contrast to models based on region proposals such as Faster-RCNN\cite{ren2016faster}, as it unifies a single network to perform detection tasks, streamlining the process. 

YOLO architecture consists of the backbone, neck, and head, which are responsible for representative feature extraction, feature enhancement, and task-specific operations such as classification and regression, respectively.
YOLOv5 achieves 50.5\% mAP on the COCO, and was soon surpassed by YOLOv8, which attains 53.9\% mAP on the same dataset.

\subsection{Overview of Detection Network}
Although YOLOv8 has achieved certain results in practical applications, there is still room for improvement in scenarios like badminton matches, where require both high real-time performance and precision.

We propose YO-CSA architecture, depicted in Fig.\ref{fig:yocsa}. Inspired by contextual transformer block (CoT)\cite{li2022contextual} and spatial group-wise enhance (SGE)\cite{li2019spatial}, we introduce contextual transformer block with 2 convolutions (CoT2f) and spatial attention-integrated neck (SANeck), which strengthen the network’s ability to extract and enhance features, particularly in terms of positional distribution, within the backbone and neck processes. Furthermore, we optimize the detection head to facilitate more efficient learning of spatial distributions. Further details on the YO-CSA architecture is provided in the following sections.

\subsection{Contextual Transformer Block with 2 Convolutions}
It is crucial to design a deep neural network with robust representative extraction capability while mitigating computational overhead and minimizing information loss. CoT2f is implemented in the backbone of YOLO aiming to enhancing the property to extract global context as well as alleviate representative information degradation. The foundational launching point for our approach is to fully exploit the contextual self-attention mechanism and Bottleneck\cite{he2016deep} recognized as an efficient architectural paradigm to boost network’s learning while reducing computational demands.

\begin{figure}[htbp]
    \centering  
    \begin{subfigure}[b]{0.31\textwidth}
        \centering
        \includegraphics[width=\textwidth]{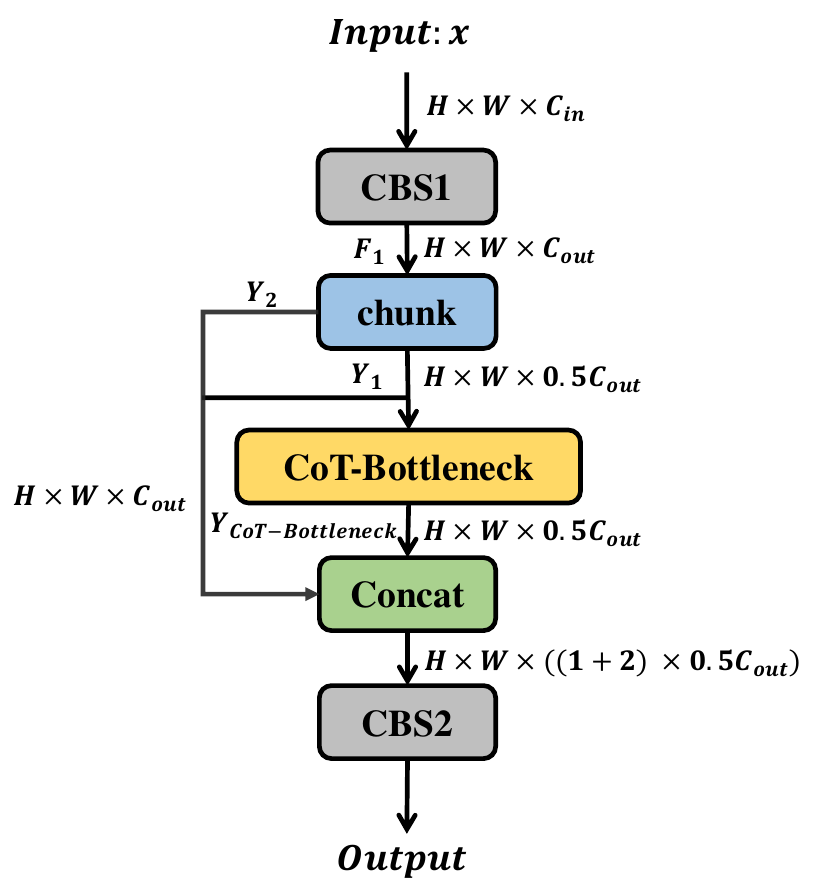}
        \caption{CoT2f}
        \label{fig:CoT2f}
    \end{subfigure}%
    \hfill
    \begin{subfigure}[b]{0.17\textwidth}
        \centering
        \includegraphics[width=\textwidth]{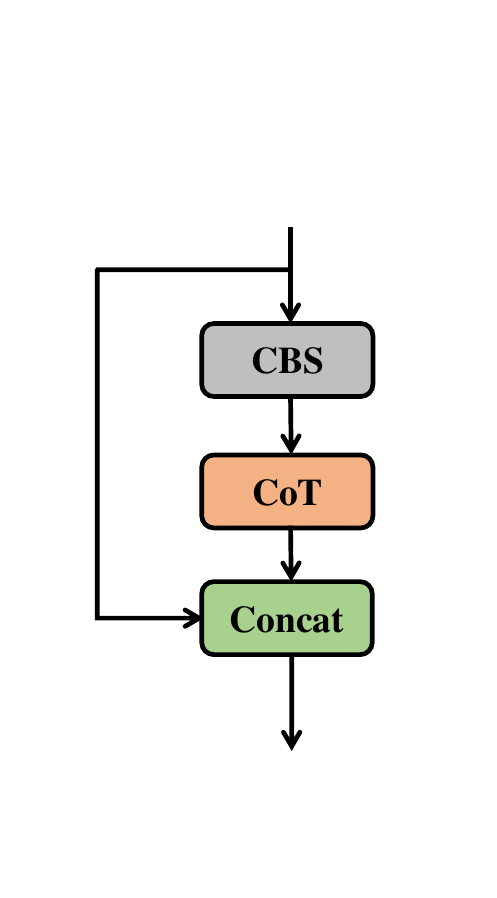}
        \caption{CoT-Bottleneck}
        \label{fig:CoT-BottleNeck}
    \end{subfigure}%
    \caption{Structure of CoT2f}
    \label{fig:cot2f}
\end{figure}

As Fig.\ref{fig:cot2f} illustrates, CoT2f comprises two convolution layers of different sizes and a CoT-Bottleneck. Initially, input X with the size of $W \times H \times {C_1}$ is fed into convolution layer 1, resulting in an output with the shape of $W \times H \times {2c}$, here $2c$ denotes the number of channels in the convolution layer 1. Subsequently, an inter-mediate product $[Y_1, Y_2]$ is obtained through a chunking operation.

CoT-Bottleneck is capable of conducting finer-grained feature extraction on ${Y_1}$ to yield $Y_\text{CoT-Bottleneck}$, whereas ${Y_2}$ is directly injected into the concatenation layer. Analogous to a residual structure, $Y_\text{CoT-Bottleneck}$ and ${Y_2}$ are concatenated, followed by processing through convolution layer 2 to facilitate the fusion of features. CoT\cite{li2022contextual}, incorporated within CoT-BottleNeck, integrates contextual information mining with a self-attention mechanism into a cohesive architecture. Unlike traditional self-attention designs, it employs $3\times3$ convolution kernel to capture contextual information, rather than relying on independent $1\times1$ kernel, which decompose the correlations between paired key-value mappings.

\subsection{Spatial Attention-Integrated Neck}
Within the context of neural networks, the target object is  composed of a sequence of sub-features, which implies that accurately identifying these sub-features enables the precise localization of the target object. Therefore, we emphasize the importance of sub-features perception and extraction in the neck of YOLO, specifically within the feature fusion process of the dual pyramid structure. To be more precise, a new dual pyramid structure is reconstructed based on SANeck, with the aim of enhancing network’s semantic extraction capability through self-attention mechanism.

\begin{figure}[h]
    \centering  
    \includegraphics[width=0.48\textwidth]{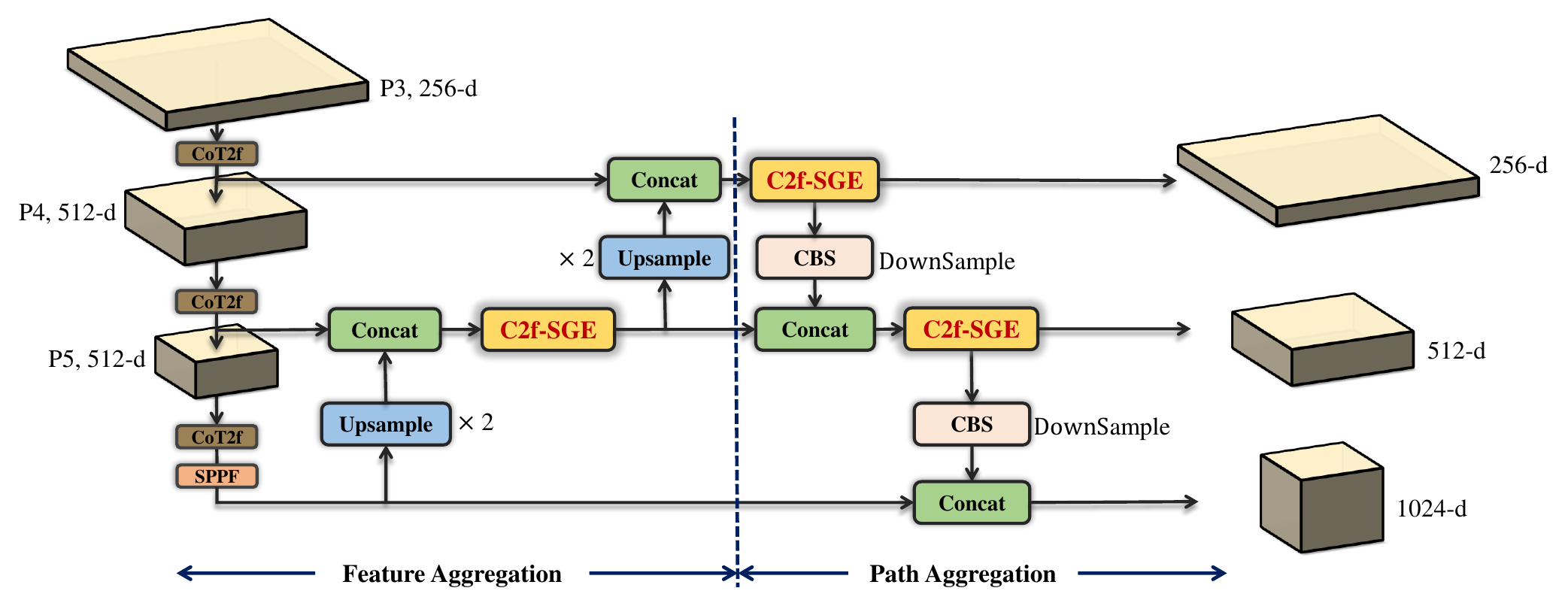}
    \caption{Spatial Attention-Integrated Neck}
    \label{fig:saneck}  
\end{figure}

For the purpose of minimizing computational overhead, SANeck appropriately references the lightweight structure SGE\cite{li2019spatial}, which draws inspiration from CapsNet\cite{sabour2017dynamic}. Inspired by the C2f structure in YOLOv8, we reconstruct neck by C2f-SGE to explicitly introduce self-attention in the neck. Fig.\ref{fig:saneck} illustrates detail of SANeck. Firstly, the output from the previous layer is fed to convolution layer 1. Then, the convolved output is chunked into two slices, one slice, ${Y_1}$, is compressed by n SGE-2f modules, while the other slice, ${Y_2}$, is directly transported to the concatenation operation.

\begin{figure}[h]
    \centering  
    \includegraphics[width=0.48\textwidth]{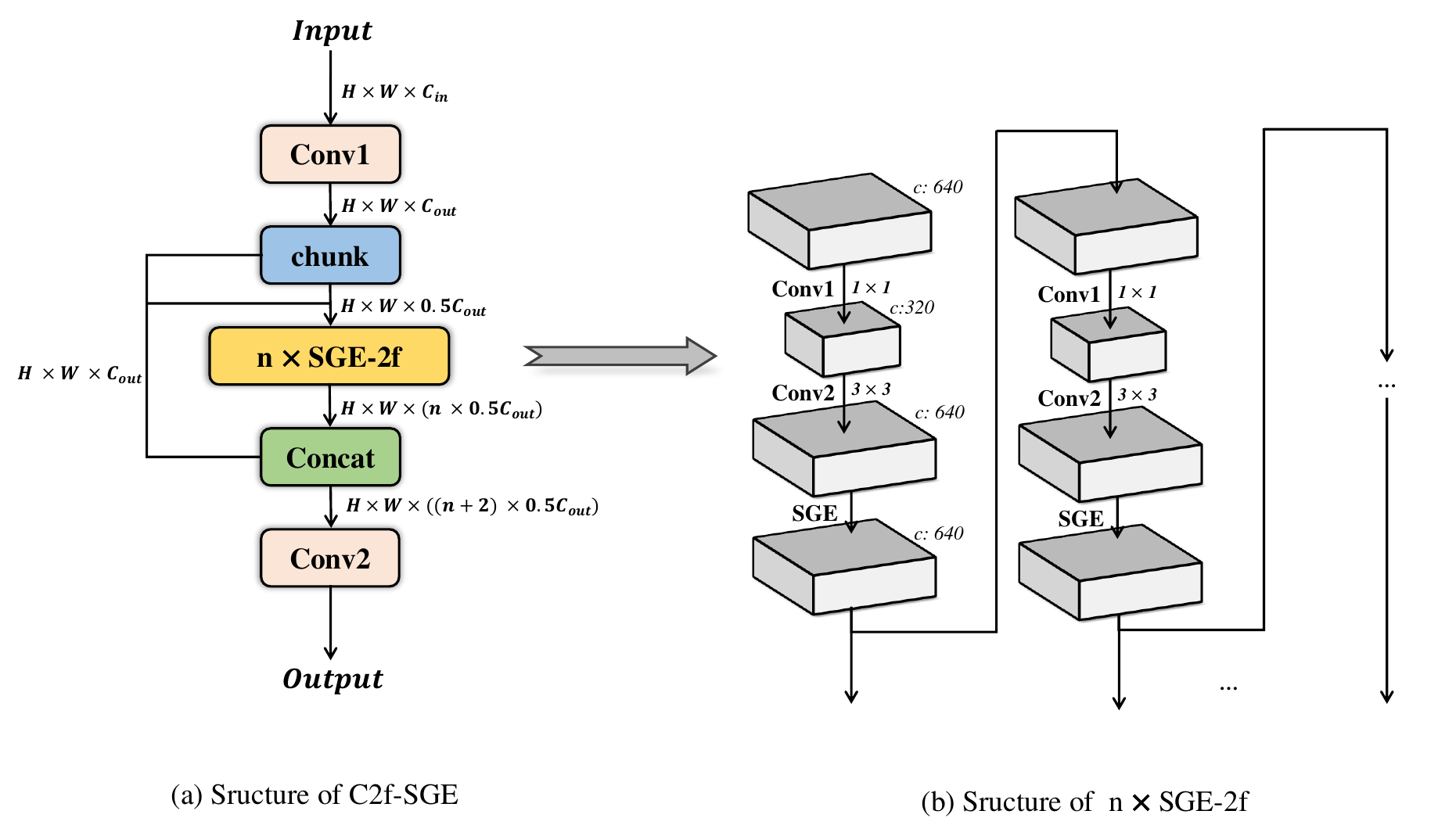}
    \caption{SGE with 2 Convolution}
    \label{fig:c2f-sge}  
\end{figure}

\subsection{Decouple head with SGE}
According to the paradigm of object detection, the backbone extracts representative features, while the downstream layers handle tasks, such as classification and bounding box regression. Thus, greater attention should be given to spatial distribution in these downstream layers.

Contrasting with the traditional detection head, the decoupled head no longer shares the parameters between classification and regression. Instead, it utilities two parallel branches to decouple the two tasks, allowing the network to learn the spatial distribution for each task independently.

As mentioned, SGE\cite{li2019spatial} highlights the similarity between local and global features, helping the network learn spatial distributions more effectively. Therefore, the decoupled head integrated with SGE outperforms the traditional version.

\begin{figure}[h]
    \centering  
    \includegraphics[width=0.48\textwidth]{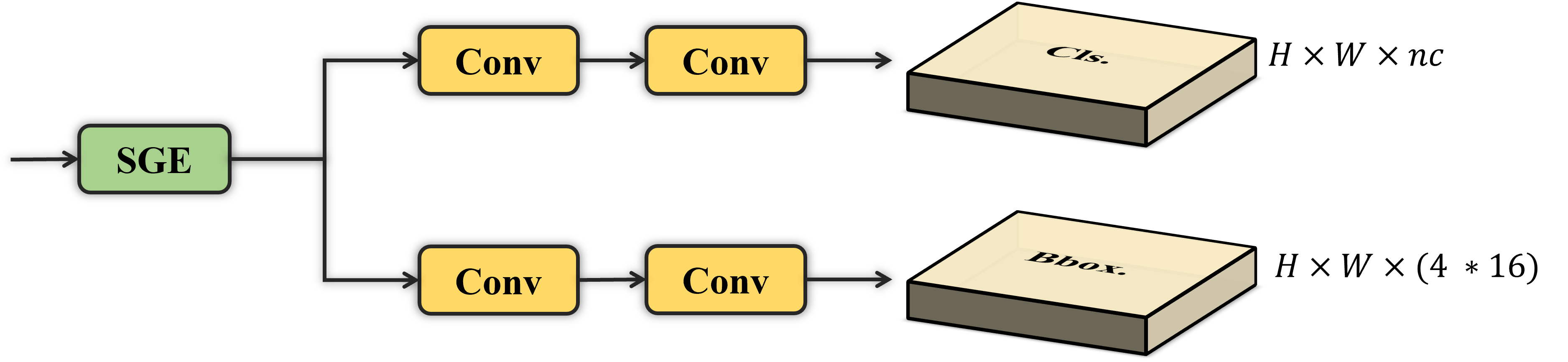}
    \caption{Decouple head with SGE}
    \label{fig:head}  
\end{figure}

As shown in Fig.\ref{fig:head}, after processed by SGE, the single branch is split into two parallel branches, each containing two 3x3 convolutions, which perform classification and regression, respectively. The output of classification and regression are [H, W, nc] (where nc denotes the number of classes) and [H, W, 64] (with 16 channels for the DFL module and 4 essential parameters for the bounding box), respectively.

\section{YO-CSA-T System Design}
Our total system can be decomposed into two main components: hardware infrastructure of stereo vision and software design. The software design integrates 2D object detection, advanced 3D tracking methodologies and the compensation module.

\subsection{Hardware Infrastructure of Stereo Vision}
To obtain 3D coordinates, cameras with 3D reconstruction capabilities are required. However, stereo cameras available on the market, like ZED, often struggle to achieve frame rates above 90 fps, which compromises the reliability of trajectory extraction for high-speed shuttlecocks. This is particularly problematic when capturing the distinct trajectory of a serve, which involves an initial descent followed by a parabolic ascent and then another descent.

\begin{figure}[h]
    \centering  
    \includegraphics[width=0.35\textwidth]{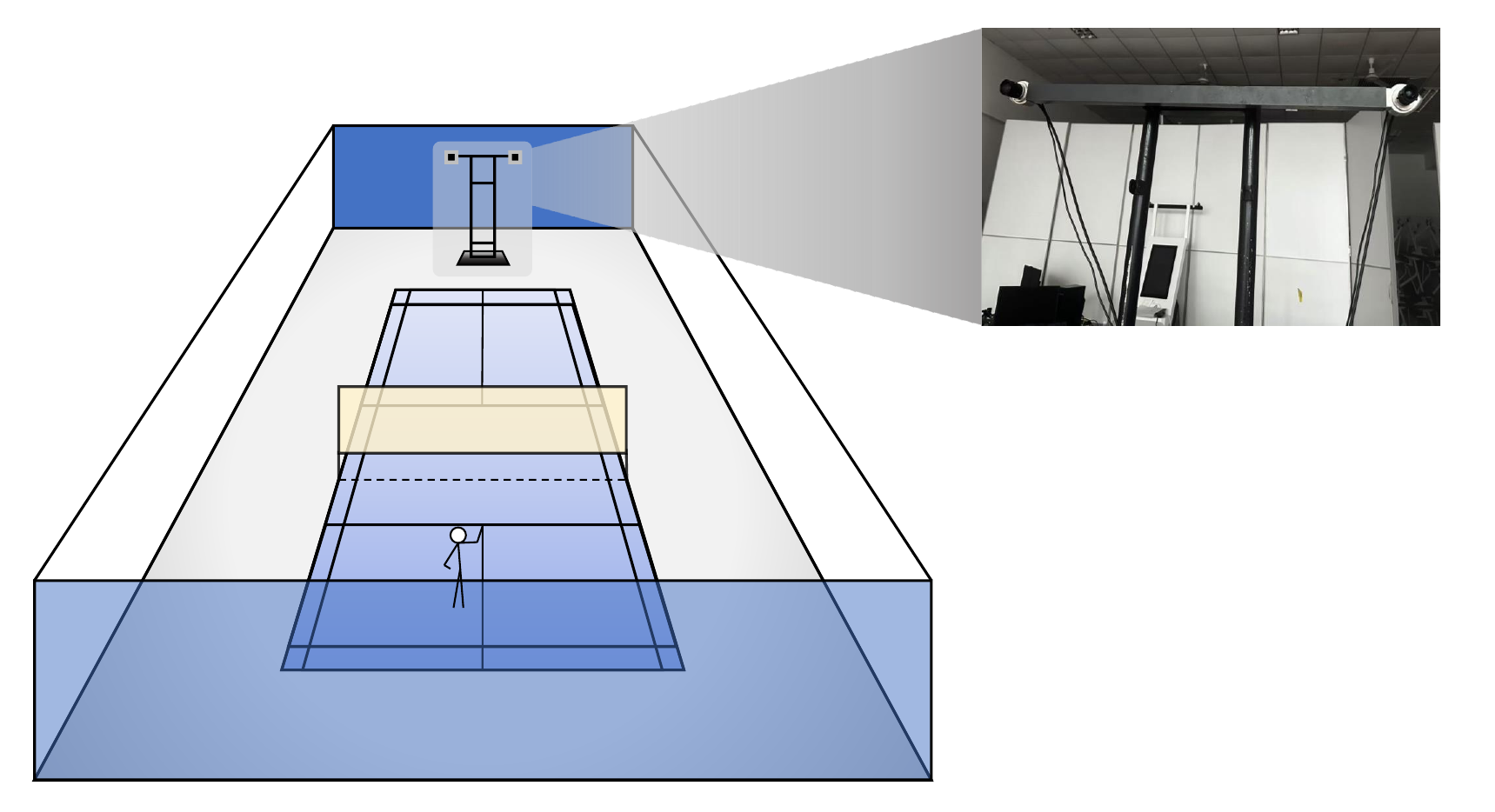}
    \caption{Schematic Diagram of a Stereo Vision Setup}
    \label{fig:setup}  
\end{figure}

To address this issue, we employ two monocular cameras (model A7200CU130, manufactured by Huairui Technology) to construct a stereo vision system. This setup enables the creation of a high-precision binocular system capable of covering the entire badminton court while maintaining accuracy. Fig.\ref{fig:setup} illustrates the placement of the cameras and their supporting structures during implementation. Specifically, the stereo cameras are positioned at the rear of the robot’s court to clearly capture the trajectory of incoming shuttlecocks. The baseline between cameras is set at 0.8 m, and the cameras are mounted at a height of approximately 1.8 m to simulate the perspective of an adult male.

\subsection{Detection Module}
 The detection network YO-CSA, core of detection module, explicitly incorporates con-textual and spatial attention mechanisms and successfully achieves performance that surpasses YOLOv8 and YOLO11. The specific structure of our detection network is presented in detail in Section \uppercase\expandafter{\romannumeral3}.

\begin{figure*}[htbp]
    \centering  
    \begin{subfigure}[b]{0.85\textwidth}
        \centering
        \includegraphics[width=\linewidth]{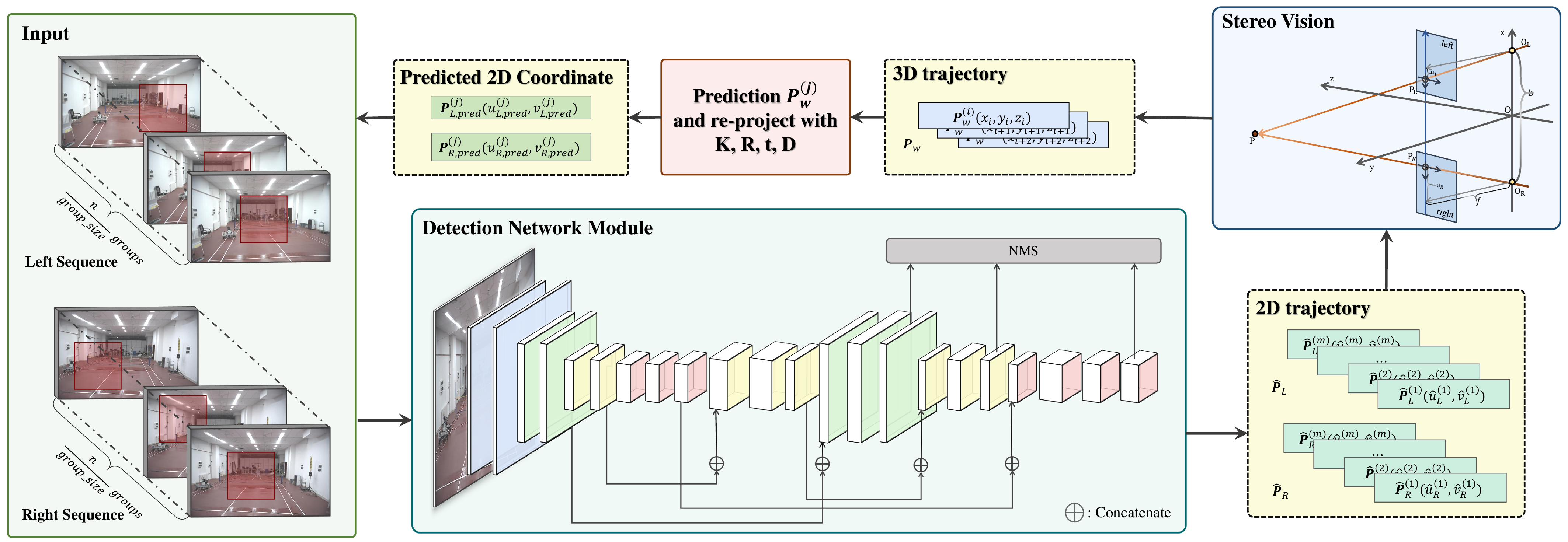}
        \caption{The process of extracting 3D trajectories from the left and right imaging planes.}
        \label{fig:dataset_scenes}
    \end{subfigure}%
    \hfill
    \begin{subfigure}[b]{0.7\textwidth}
        \centering
        \includegraphics[width=\linewidth]{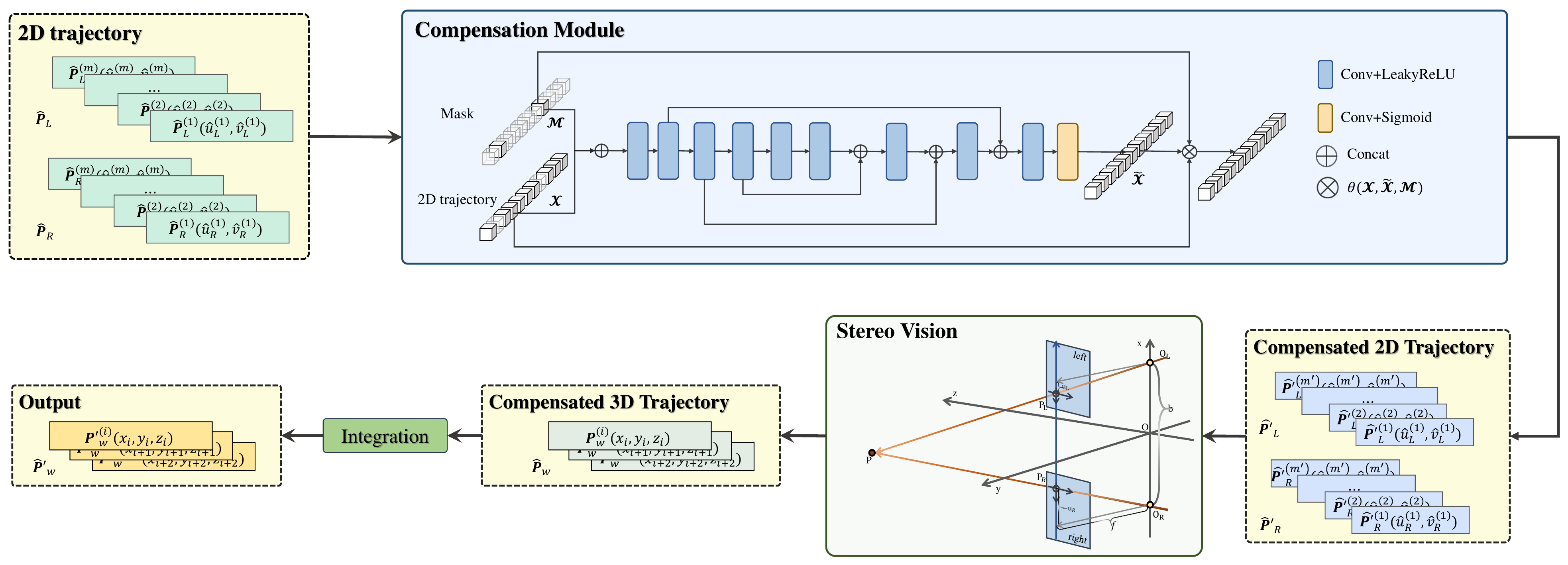}
        \caption{Compensation process}
        \label{fig:effects_of_aug}
    \end{subfigure}%
\caption{System workflow: (a) The process of extracting 3D trajectories from the left and right imaging planes, (b) Compensation process.}
\label{fig:workflow}
\end{figure*}

\subsection{Comprehensive Tracking Workflow}
Our pipeline not only detects the shuttlecock in the current frame, but also predicts its future trajectory based on historical information, thereby incorporating reasonable constraints to enhances the plausibility of the detection range. Throughout the process, we implement constraints to ensure trajectory consistency, such as region of interest (ROI) and threshold settings $\varepsilon_1$, $\varepsilon_2$, $\varepsilon_3$ evaluating the plausibility of coordinates.

Fig.\ref{fig:workflow} displays our entire tracking pipeline based on 2 image sequences, denoted as $\mathbf{I}_L = [{\mathbf{I}^{(1)}_{L}}, …, {\mathbf{I}^{(n)}_{L}}]$, $\mathbf{I}_R = [{\mathbf{I}^{(1)}_{R}}, …, {\mathbf{I}^{(n)}_{R}}]$, captured by the left and right monocular cameras respectively, where each image in the sequences corresponds to a specific timestamp, ensuring temporal alignment between the left and right camera frames.

In accordance with the rules of the game, the shuttlecock's initial position is typically located at the center of the frame. So we center a $640 \times 640$ ROI area on ${\mathbf{I}^{(i)}_{L}}$ and ${\mathbf{I}^{(i)}_{R}}$, which not only focuses the detection process on a smaller region but also alleviates the computational burden on the detection network.

The detection network YO-CSA performs parallel detection tasks on the paired images  ${\mathbf{I}^{(i)}_{L}}$, ${\mathbf{I}^{(i)}_{R}}$. A detection is considered valid, and tracking is initiated only when both ${\mathbf{I}^{(i)}_{L}}$ and ${\mathbf{I}^{(i)}_{R}}$ successfully detect the shuttlecock at $\mathbf{P}^{(i)}_{L}(u^{(i)}_L, v^{(i)}_L)$ and $\mathbf{P}^{(i)}_{R}(u^{(i)}_R, v^{(i)}_R)$, and their displacement (especially along the vertical axis) is within the predefined threshold $\varepsilon_1$. 

Simultaneously, the detected 2D positions $[\mathbf{P}^{(i)}_{L}, \mathbf{P}^{(i)}_{R}]$ are fed into the stereo matching module to obtain the corresponding 3D coordinates. In 3D space, we perform timestamp-based fitting along all three axes on the trajectory sequence, making a prior prediction $\mathbf{P}^{(j)}_{w}$ of the shuttlecock’s trajectory. Every 10 frames, this 3D prior prediction is projected onto the left and right image planes as $\mathbf{P}^{(j)}_{L,pred}$, $\mathbf{P}^{(j)}_{R,pred}$, providing feedback to adjust the ROI displacement error. 

To ensure trajectory consistency, we also perform a plausibility check on the predicted future coordinates based on the threshold $\varepsilon_2$.

\subsection{Compensation Module}

Inspired by TrackNetv3\cite{chen2023tracknetv3}, we add a compensation module as an auxiliary branch. We use the 2D trajectory $\textbf{\^{P}}_{L}$, $\textbf{\^{P}}_{R}$ obtained from the detection module and a corresponding trajectory mask as input, employing a compensation network for interpolation. This network compensates for missing frames that were either missed during detection or discarded due to violations of spatiotemporal constraints. The compensation network is a U-shaped network\cite{ronneberger2015u} based on 1D convolution operations, leveraging an encoder-decoder architecture to extract and integrate both shallow spatial information and deep semantic features from the 2D trajectory.

We reintroduce the compensated trajectory $\hat{\mathbf{P}^{'}}_L$, $\hat{\mathbf{P}^{'}}_R$ into the stereo vision module to obtain the complete 3D trajectory $\hat{\mathbf{P}}_w$. Based on spline interpolation, we generated another complete trajectory $\tilde{\mathbf{P}}_w$ from the trajectory ${\mathbf{P}}_w$. Subsequently, $\tilde{\mathbf{P}}_w$, the compensated trajectory $\hat{\mathbf{P}}_w$ and mask $\mathbf{M}$ are integrated through function $\theta$ to obtain fully completed trajectory $\hat{\mathbf{P}^{'}}_w$ as \eqref{eq:theta}.
\begin{footnotesize}
\begin{equation}
\label{eq:theta}
\hat{\mathbf{P}^{'}}^{(i)}_{w}= \\
\begin{cases} 
\mathbf{P}^{(i)}_w, & \text{if } M_{i} = 1 \\
\alpha \mathbf{\hat{P}}^{(i)}_w - \alpha \mathbf{\tilde{P}}^{(i)}_w, & \text{if } M_{i}=0 \text{ and } ||\mathbf{\hat{P}}^{(i)}_w - \mathbf{P}^{(j)}_w|| \leq \varepsilon_3
\end{cases}
\end{equation}
\end{footnotesize}

\section{Experiments and Results}
\subsection{Dataset}

\begin{figure}[htbp]
    \centering  
    \begin{subfigure}[b]{0.23\textwidth}
        \centering
        \includegraphics[width=\textwidth]{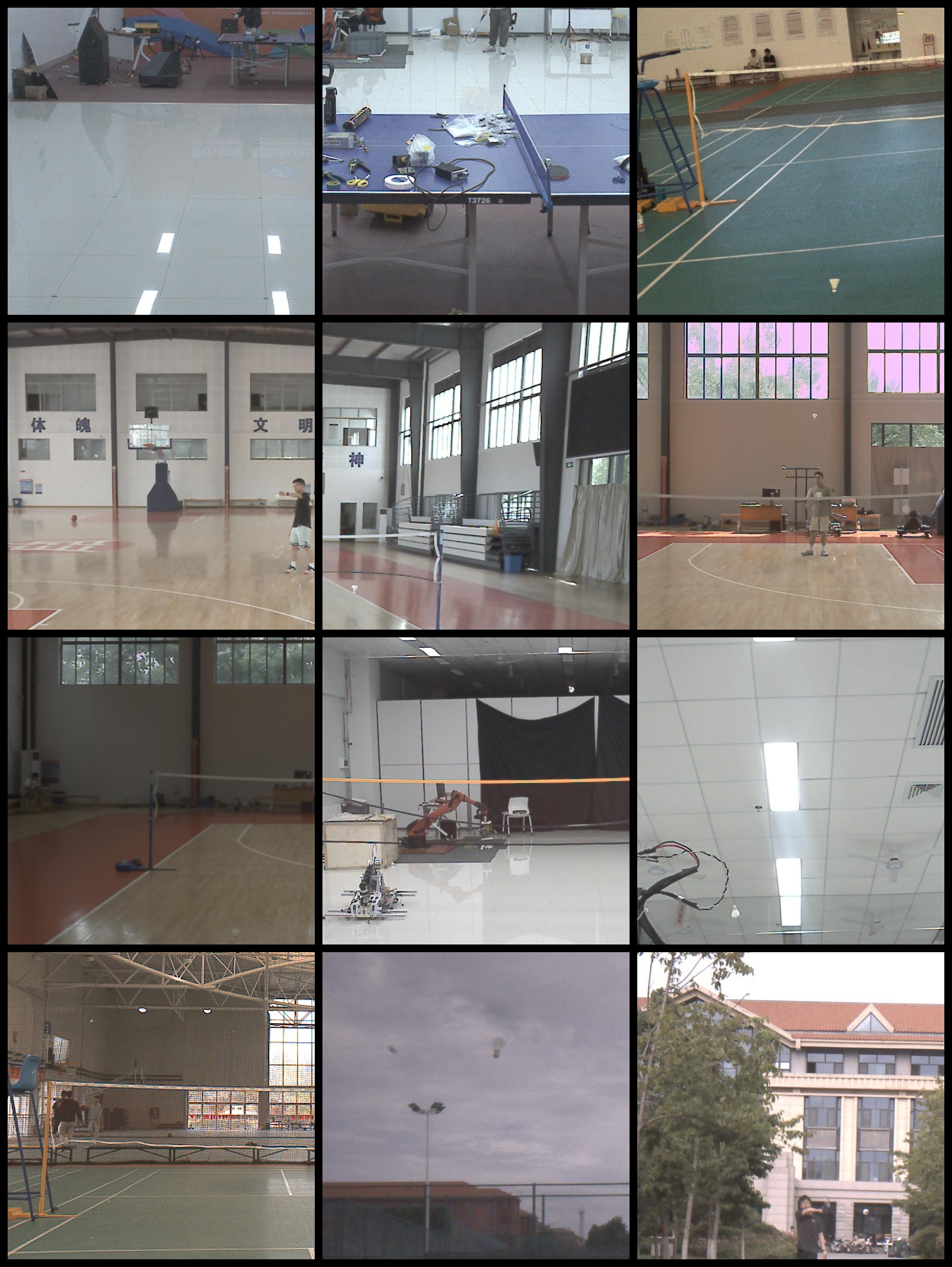}
        \caption{Different Scenes from the Dataset}
        \label{fig:dataset_scenes}
    \end{subfigure}%
    \hfill
    \begin{subfigure}[b]{0.23\textwidth}
        \centering
        \includegraphics[width=\textwidth]{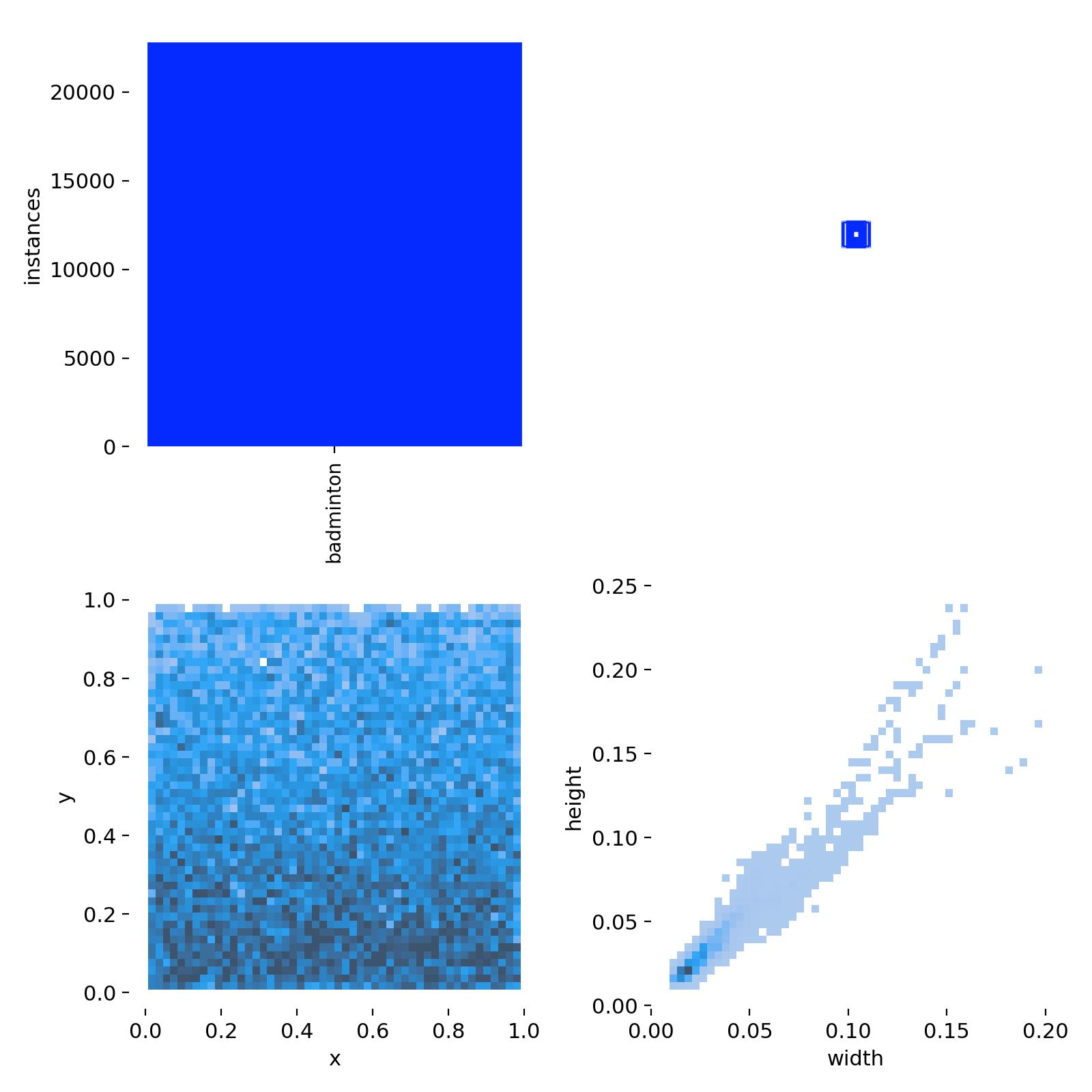}
        \caption{Shuttlecock Distribution in the Training Set}
        \label{fig:shuttlecock_distribution}
    \end{subfigure}%
\caption{Representative scenes in our dataset.}
\end{figure}

Based on the custom-built 3D vision system, we collect badminton rally data from 10 distinct venues or environments, encompassing various angles and natural lighting conditions, totaling 32,539 images. Random $640 \times 640$ pixel regions were sampled from the original images, with the condition that the shuttlecock must be present within the cropped area. Specifically, the shuttlecock is not necessarily centered within the cropping region but is randomly distributed across the $640 \times 640$ area. Fig. \ref{fig:dataset_scenes} illustrates a subset of our dataset.

During actual training, to enhance the generalization and robustness of the network, we also applies data augmentation techniques, including adjustments to the HSV hue, saturation, and value, as well as geometric transformations such as rotation, translation, scaling, and flipping.

The training set label description for our dataset is shown in Fig.\ref{fig:shuttlecock_distribution}. As observed from the figure, the shape of the shuttlecock in the dataset predominantly appears as a nearly square-like rectangle, which aligns with the fact that the standard dimensions of a shuttlecock are 68-78mm in height and 58-68mm in diameter.

\subsection{Detection Experiments}
Aside from enhancing the data augmentation techniques, we retain the original parameters of YOLOv8 and train the model for 300 epochs on our custom dataset.

\begin{figure}[h]
    \centering  
    \includegraphics[width=0.35\textwidth]{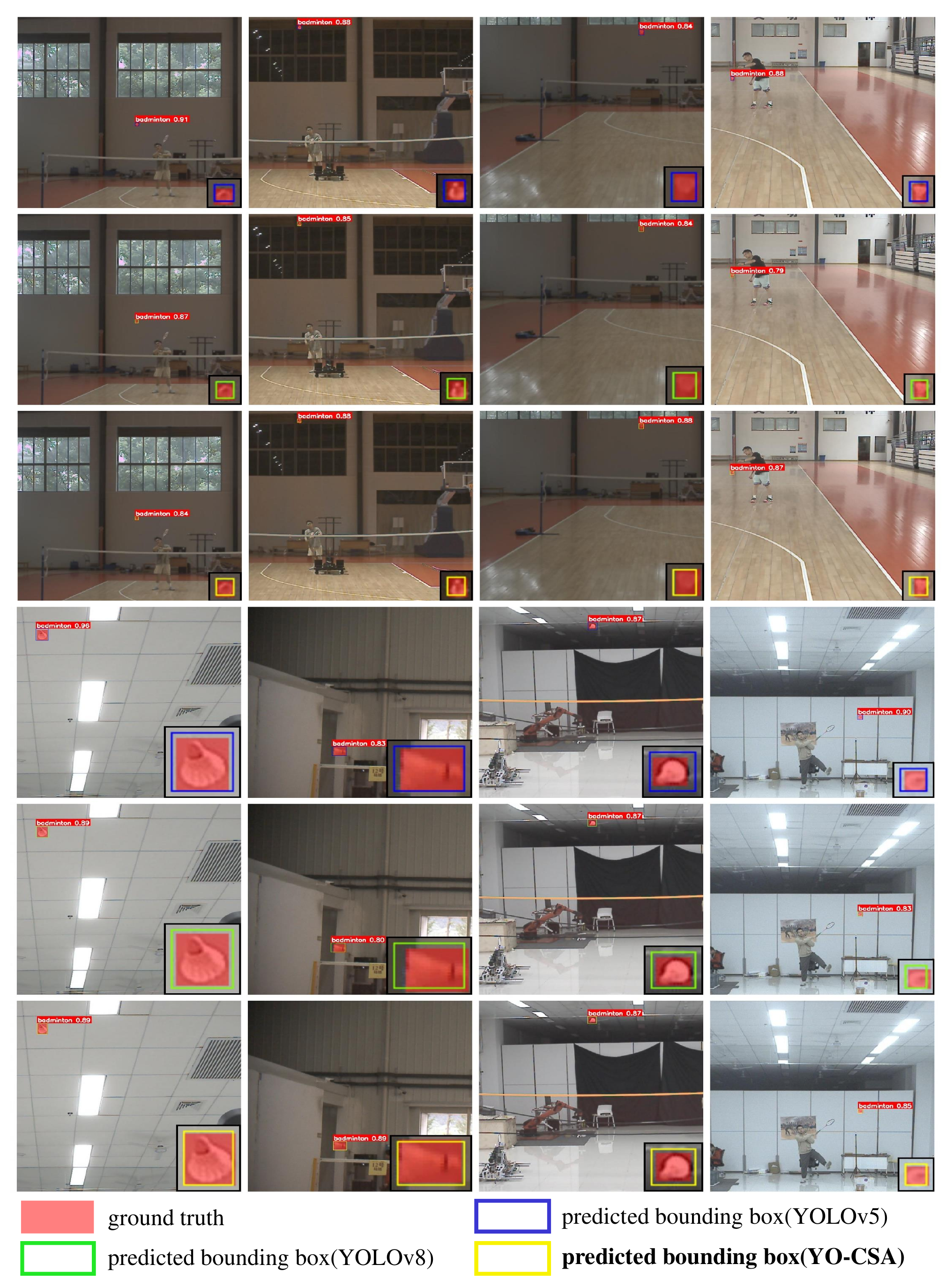}
    \caption{Detection results including labels and confidence scores. Each image's detected bounding boxes are magnified and displayed in the bottom-right corner with black borders.}
    \label{fig:vis_detection}  
\end{figure}

\begin{table}[ht]
    \centering
    \captionsetup{justification=centering, labelsep=none}
    \caption{\scriptsize \\ COMPARISON EXPERIMENTS WITH OTHER MAINSTREAM MODELS}
    \label{tbl:compared_with_other_models}
    \resizebox{0.48\textwidth}{!}{
        \begin{tabular}{c|ccccccc}
        \Xhline{2pt}
        Model        & \multicolumn{1}{c}{mAP@0.5(\%)} & \multicolumn{1}{c}{mAP@0.75(\%)} & \multicolumn{1}{c}{mAP@0.5:0.95(\%)} & \multicolumn{1}{c}{Recall(\%)} & \multicolumn{1}{c}{FPS} \\ \Xhline{1pt}
        Faster R-CNN      &98.30 &78.29 &58.52 &87.10 &229.6 \\
        RetinaNet      &93.87 &72.76 &59.57 &74.09 &222.4 \\
        YOLOv5s      &98.49 &76.06 &65.59  &95.28 &\textbf{285.7} \\
        YOLOv8s      &99.38 &82.67 &71.50  &98.62&250.0 \\ 
        YOLO11s      &99.37 &86.94 &71.56  &98.25 &238.1 \\ 
        \textbf{YO-CSA(Ours)} &\textbf{99.43} &\textbf{90.43} &\textbf{74.00}  &\textbf{99.02} &243.9 \\
        \Xhline{2pt}
        \end{tabular}
    }
\end{table}

Our detection network achieves 74.00\% mAP@0.5:0.95 on our dataset, while YOLOv5s achieving 65.59\% and YOLOv8s achieving 71.50\%. Table \ref{tbl:compared_with_other_models} presents the detection results of the YO-CSA algorithm alongside other mainstream detection models. Fig.\ref{fig:vis_detection} shows the visual results of the YO-CSA on our dataset, demonstrating a high level of accuracy.

The accuracy of boxes directly influences the precision of trajectory prediction. Therefore, in addition to focusing on precision, we place particular emphasis on the mAP metric. Considering that mAP@0.5 does not fully capture the improvements, we have also highlighted mAP@0.75. Our detection network achieves 90.43\% mAP@0.75, outperforming YOLOv8s by 7.76\% and YOLO11s by 3.49\%.

To validate the effectiveness of our optimizations, we conducts a series of ablation experiments, with results presented in Table \ref{tbl:ablation_exps}. The data demonstrate that the improved YO-CSA achieves the best performance while even reducing the network's GFLOPs.

\subsection{Tracking Experiments}
In order to assess the effectiveness of our approach, we collect 12 video clips of serve and hit actions, each captured at 160fps, using our custom-built vision system. Although the YO-CSA design already incorporates a lightweight structure, we aim to further optimize the detection speed to meet the demands of high-level competitions. Therefore, we accelerate the model using ONNX. Table \ref{tbl:lightweight_preformances} presents the detection performance of the accelerated YO-CSA. Despite maintaining a detection accuracy of 73.94\% mAP, we are able to improve the detection speed to 5.82ms per frame, achieving a speedup of 12.74\%.

In the whole tracking process, in addition to defining ROI to alleviate the detection module, we further impose constraints in the 3D visual space on the 2D detections.

We compare the performance of 4 strategies in Table \ref{tbl:system_performance}: (a) performing object detection directly on the 2D image sequences, obtaining paired left and right view coordinate sequences, and then conducting 3D vision matching;
(b) restricting fixed 640x640 ROI regions on the left and right views;
(c) based on (b), utilizing historical trajectory information to predict the next 3D position in 3D space and projecting it onto the left and right views to update the ROI; (d) based on (c), adopting compensation module to optimize the trajectory.

\begin{table}[ht]
    \centering
    \captionsetup{justification=centering, labelsep=none}
    \caption{\scriptsize \\ ABLATION EXPERIMENTS}
    \label{tbl:ablation_exps}
    \resizebox{0.48\textwidth}{!}{
        \begin{tabular}{c|ccccccc}
        \Xhline{2pt}
        Model        & \multicolumn{1}{c}{mAP@0.5(\%)} & \multicolumn{1}{c}{mAP@0.75(\%)} & \multicolumn{1}{c}{mAP@0.5:0.95(\%)} & \multicolumn{1}{c}{Recall(\%)}  & \multicolumn{1}{c}{GFLOPs} \\ \Xhline{1pt}
        YOLOv8s     &99.38 &82.67 &71.50  &98.62 &23.6 \\ 
        YOLOv8+ab   &99.40 &89.21 &73.32  &98.91 &20.9\\ 
        YOLOv8+ac   &99.40 &89.37 &73.56  &98.40  &20.9 \\ 
        YOLOv8+bc   &99.42 &88.37 &73.54  &98.97 &23.6 \\ 
         \textbf{YOLOv8+abc(Ours)} &\textbf{99.43} &\textbf{90.43} &\textbf{74.00}  &\textbf{99.02} &\textbf{20.9} \\ 
        \Xhline{2pt}
        \end{tabular}
    }
\end{table}

\begin{table}[ht]
    \centering
    \captionsetup{justification=centering, labelsep=none}
    \caption{\scriptsize \\ LIGHTWEIGHT PERFORMANCE COMPARISON EXPERIMENTS}
    \label{tbl:lightweight_preformances}
    \resizebox{0.48\textwidth}{!}{
        \begin{tabular}{c|ccccccc}
        \Xhline{2pt}
        Model        & \multicolumn{1}{c}{mAP@0.5(\%)} & \multicolumn{1}{c}{mAP@0.75(\%)} & \multicolumn{1}{c}{mAP@0.5:0.95(\%)} & \multicolumn{1}{c}{Recall(\%)} & \multicolumn{1}{c}{Inference Time(ms)} \\ \Xhline{1pt}
        YOLOv8s      &99.38 &82.67 &71.50  &98.62 &6.84 \\ 
        YOLOv8s(accelerated)      &99.34 &82.65 &71.27  &98.55 &5.93 \\  
        YOLO11s      &99.37 &86.94 &71.56  &98.25 &7.03 \\ 
        YOLO11s(accelerated)      &99.36 &86.45 &71.23  &98.17 &6.66 \\ 
        YO-CSA(Ours)      &\textbf{99.43} &\textbf{90.43} &\textbf{74.00}  &\textbf{99.02} &6.67 \\ 
        \textbf{YO-CSA(Ours, accelerated)} &99.42 &90.41 &73.94  &99.01 &\textbf{5.82} \\ 
        \Xhline{2pt}
        \end{tabular}
    }
\end{table}

Since obtaining the ground truth of 3D trajectories is highly costly, we evaluate 4 strategies by computing the smoothness of the trajectory based on velocity and acceleration, namely velocity smoothness ($S_v$) ans acceleration smoothness ($S_a$), as well as the average centroid shift of the trajectory ($C_{avg}$). For the 3D trajectory ${\textbf{P}_w(x^{(i)}_{w},y^{(i)}_{w},z^{(i)}_{w})}$:

\begin{footnotesize}
\begin{equation}
\textbf{v}_{i}=\frac{\textbf{p}^{(i+1)}_{w}-\textbf{p}^{(i)}_w}{t_{i+1}- t_i}, \quad\textbf{a}_{i}=\frac{\textbf{v}_{i+1}-\textbf{v}_{i}}{t_{i+1}- t_i}
\end{equation}
\begin{equation}
S_v=\frac{1}{n-2}\sum^{n-1}_{i=2}{\Delta{v_i}^2},\quad
\Delta{v_i}=||\textbf{v}_i-\textbf{v}_{i-1}||
\end{equation}
\begin{equation}
S_a=\frac{1}{n-2}\sum^{n-1}_{i=2}{\Delta{a_i}^2},\quad
\Delta{a_i}=||{\frac{\textbf{v}_i-\textbf{v}_{i-1}}{(t_{i+1}-t_{i-1})/2}}||
\end{equation}
\begin{equation}
C_{avg}=\frac{1}{n-1}\sum^{n-1}_{i=1}\Delta{{C}_i},
\end{equation}
\begin{equation}
\Delta{C_i}=\sqrt{(\textbf{x}^{(i+1)}_{w}-\textbf{x}^{(i)}_{w})^2+(\textbf{y}^{(i+1)}_{w}-\textbf{y}^{(i)}_{w})^2+(\textbf{z}^{(i+1)}_{w}-\textbf{z}^{(i)}_{w}}
\end{equation}
\end{footnotesize}

\begin{figure}[htbp]
    \centering  
    \begin{subfigure}[b]{0.48\textwidth}
        \centering
        \includegraphics[width=\textwidth]{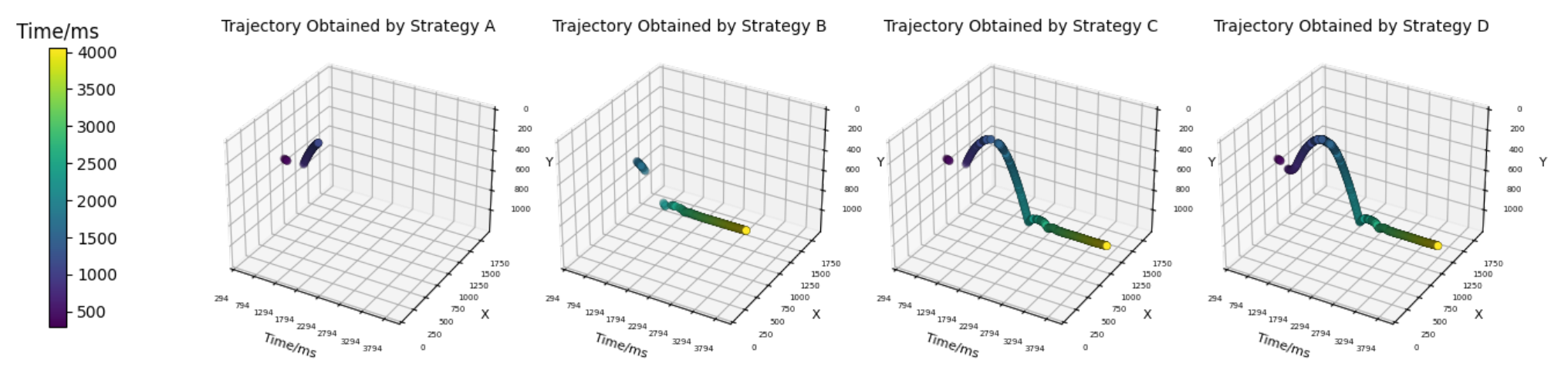}
        \caption{Comparison of 2D trajectories on the left imaging plane}
        \label{fig:3d_left}
    \end{subfigure}%
    \hfill
    \begin{subfigure}[b]{0.48\textwidth}
        \centering
        \includegraphics[width=\textwidth]{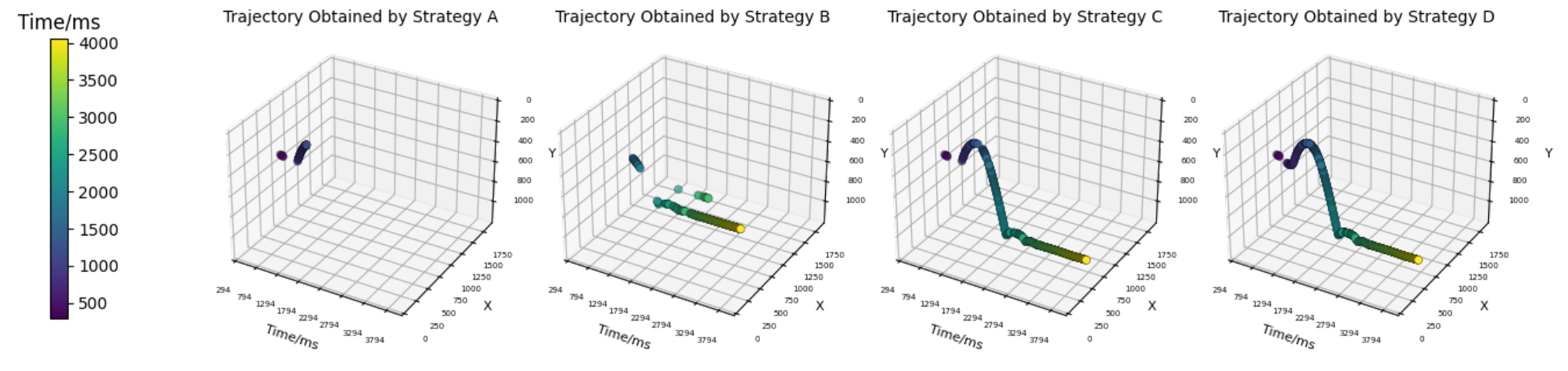}
        \caption{Comparison of 2D trajectories on the right imaging plane}
        \label{fig:3d_right}
    \end{subfigure}%
    \hfill
    \begin{subfigure}[b]{0.48\textwidth}
        \centering
        \includegraphics[width=\textwidth]{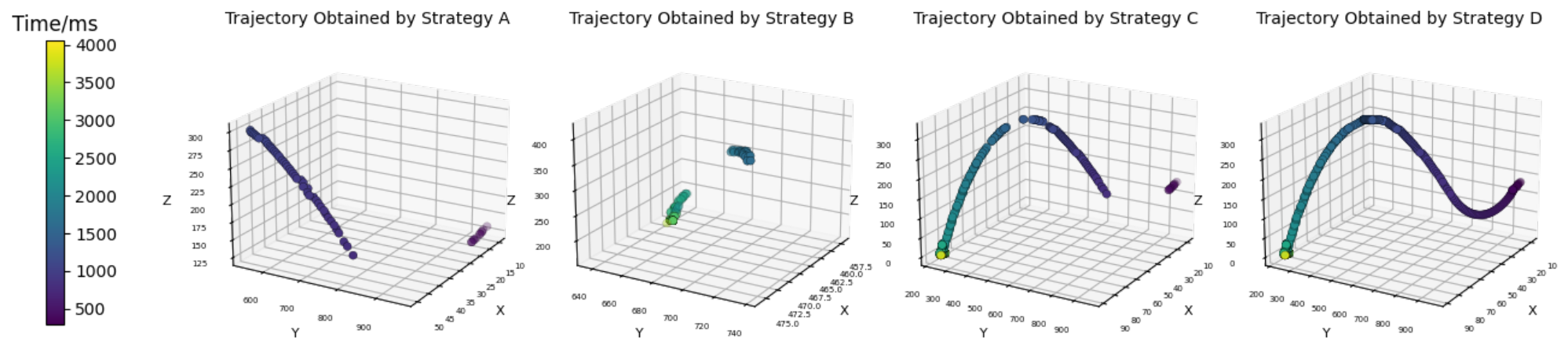}
        \caption{Comparison of 3D trajectories}
        \label{fig:3d}
    \end{subfigure}%
\caption{Comparison of trajectories extracted using four different strategies.}
\label{fig:four strategies}
\end{figure}

We assume that smaller smoothness and centroid shifts indicate trajectories that are closer to reality. As displayed in Table \ref{tbl:system_performance}, Strategy D performs the best. Fig.\ref{fig:four strategies} illustrates the results of four strategies .

\begin{table}[ht]
    \centering
    \captionsetup{justification=centering, labelsep=none}
    \caption{\scriptsize \\ SYSTEM PERFORMANCE WITH DIFFERENT STRATEGIES}
    \label{tbl:system_performance}
    \resizebox{0.48\textwidth}{!}{
        \begin{tabular}{c|ccccccc}
        \Xhline{2pt}
        Strategy & \multicolumn{1}{c}{Acceleration(\%)} & \multicolumn{1}{c}{Velocity Smoothness} & \multicolumn{1}{c}{Acceleration Smoothness} & \multicolumn{1}{c}{Average Centroid Shift(cm)} & \multicolumn{1}{c}{FPS} \\ \Xhline{1pt}
        Strategy A  &27.86 &$>$100 &$>$1e7 &19.10 &85.7 \\ 
        Strategy B  &30.38 &6.91 &1.13e6 &5.76 &90.9 \\ 
        Strategy C &81.28 &3.92 &2.77e5 &2.35 &\textbf{133.3} \\  
        \textbf{Strategy D} &\textbf{100.00} &\textbf{2.31} &\textbf{2.25e5} &\textbf{2.35} &- \\  
        \Xhline{2pt}
        \end{tabular}
    }
\end{table}

\section{CONCLUSION AND FUTURE WORK}
We construct YO-CSA, a network built upon YOLOv8s, leveraging spatial and contextual attention mechanisms to achieve a substantial improvement in detection performance. Additionally, we introduce a multi-dimensional spatiotemporal constraint strategy and design a real-time system for the precise extraction of the shuttlecock’s 3D trajectory based on YO-CSA. Experimental results demonstrate that our system can extract the shuttlecock's trajectory with high accuracy and in real-time. Building upon the current foundation, we intend to further investigate the practical implementation of human-machine competitions in future work.

\addtolength{\textheight}{-12cm}   





\end{document}